\documentclass{Interspeech2024}

\interspeechcameraready

\title{Soft Language Identification for Language-Agnostic Many-to-One End-to-End Speech Translation}

\name[]{Peidong}{Wang}
\name[]{Jian}{Xue}
\name[]{Jinyu}{Li}
\name[]{Junkun}{Chen}
\name[]{Aswin Shanmugam}{Subramanian}

\address{
  Microsoft, USA}
\email{\{peidongwang, jian.xue, jinyli, junkunchen, aswins\}@microsoft.com}


\begin{document}

\maketitle

\begin{abstract}
Language-agnostic many-to-one end-to-end speech translation models can convert audio signals from different source languages into text in a target language. These models do not need source language identification, which improves user experience. In some cases, the input language can be given or estimated. Our goal is to use this additional language information while preserving the quality of the other languages. We accomplish this by introducing a simple and effective linear input network. The linear input network is initialized as an identity matrix, which ensures that the model can perform as well as, or better than, the original model. Experimental results show that the proposed method can successfully enhance the specified language, while keeping the language-agnostic ability of the many-to-one ST models.
\end{abstract}

\section{Introduction}
\label{sec:intro}
Speech translation (ST) aims to transform speech signals in one language into text in another language.
The traditional way to achieve this is to use automatic speech recognition (ASR) followed by text-based machine translation (MT) . This method has some drawbacks. First, MT may be affected by ASR errors. Second, the system cannot make full use of the speech information. For many-to-one ST, this method becomes even more challenging, as each source language may require a different ASR model and a corresponding MT model.

Unlike cascaded systems that use separate ASR and MT models, end-to-end (E2E) ST is a method that directly translates speech features to texts using a single model \cite{li2022recent,vila2018end,sperber2020speech}. 
Similarly to E2E ASR, there may be two main types of E2E ST models: attention-based E2E encoder-decoder models (AED) \cite{vila2018end,sperber2020speech,wang2019token,wang2019large,wang21t_interspeech} and neural transducer-based models \cite{Graves-RNNSeqTransduction,prabhavalkar2017asr,sainath2020asr,li2020asr,saon2021asr}. 
An E2E AED ST model was first proposed by B{\'e}rard \emph{et al.} \cite{Berard2016ST} for a small French-English corpus. A similar model was used by Weiss \emph{et al.} \cite{weiss2017sequence} for the Fisher Callhome Spanish-English task and it surpassed the cascaded system on the Fisher test set. B{\'e}rard \emph{et al.} \cite{Berard2018ST} also used AED-based models for a large-scale E2E ST task. Recently, Radford \emph{et al.} applied weakly supervised multi-task learning for Whisper \cite{radford2023robust}. AED models are usually run in an offline mode.
Another type of common E2E models, neural transducer, may be more appropriate for streaming ST and has become popular recently \cite{papi2023token,yang2023diarist,papi2023leveraging}. Xue \emph{et al.} demonstrated that neural transducer, which was often used for ASR tasks, can be effectively applied to streaming ST \cite{xue2022large}. Wang \emph{et al.} showed that neural transducers can perform many-to-many streaming ASR and ST with a single model \cite{wang2022lamassu}. Liu \emph{et al.} proposed cross attention augmented transducer networks for streaming ST \cite{liu2021caat}. Tang \emph{et al.} conducted excellent research combining transducer and attention-based encoder-decoder architectures \cite{tang2023hybrid}.



A major advantage of E2E ST is that it can handle many-to-one ST with ease. A streaming multilingual and language-agnostic speech recognition and translation model using neural transducers (LAMASSU) \cite{wang2022lamassu} showed that by simply combining the audio data from various languages, the model can perform both German (DE) to English (EN) and EN to EN tasks. It was then expanded to 25 languages to EN many-to-one ST tasks \cite{xue2023weakly} and achieved competitive results. For AED based models, Whisper \cite{radford2023robust} confirmed the effectiveness of E2E ST in many-to-one ST scenarios.

Many-to-one E2E ST models usually work in a language-agnostic way, where the source languages are not required to be specified by the user. However, in many situations, language identification (LID) can be given or estimated. In these cases, we aim to enhance the performance of the many-to-one E2E ST models for the given language while preserving most of the performance for the other languages. Most prior works incorporate the LID by adding one-hot vectors of LID to the model. If the user selects an incorrect source LID, or the LID detection model is unreliable, the ST results are adversely affected, resulting in poor user experiences. This introduces a new task which we denote as soft LID for language-agnostic many-to-one E2E ST. Our goal is to leverage the source LID information while preserving the language-agnostic character and maintaining the translation quality for unspecified languages in our ST models.



\begin{figure*}[htb]
\begin{minipage}[b]{.48\linewidth}
  \centering
  \centerline{\includegraphics[width=5.5cm]{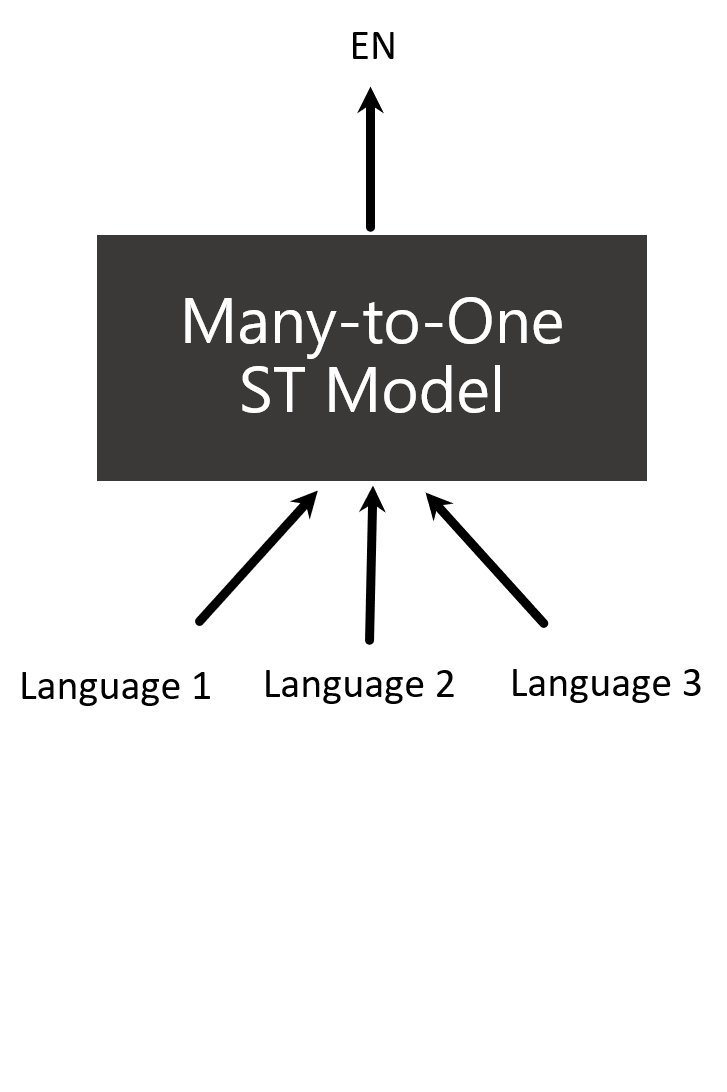}}
  \centerline{(a) A many-to-one ST model}\medskip
\end{minipage}
\hfill
\begin{minipage}[b]{0.48\linewidth}
  \centering
  \centerline{\includegraphics[width=6.2cm]{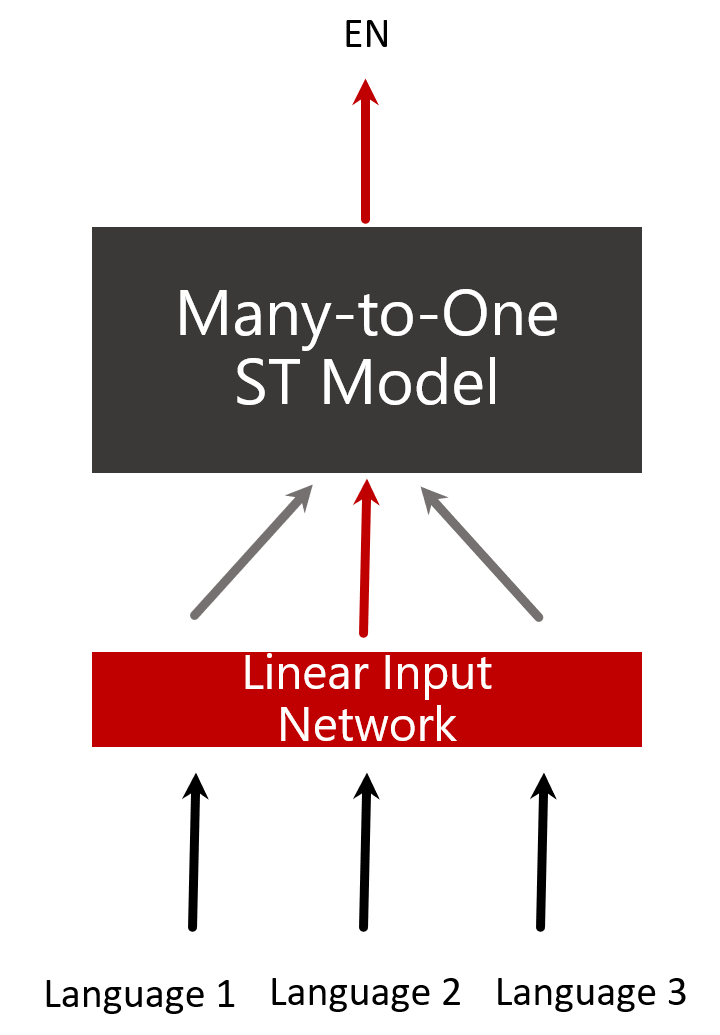}}
  \centerline{(b) A many-to-one ST model with LIN for language 2}\medskip
\end{minipage}
\caption{Illustrations of a many-to-one ST model and the proposed soft LID method using a linear input network for input language 2. For each input language, we use a different LIN layer.}
\label{fig:comparison_lin}
\end{figure*}


In this paper, we propose a solution to this problem by employing a widely used technique in the field of speaker adaptation, linear input network (LIN). Speaker adaptation is the process of reducing the discrepancy between the training and test data distributions due to speaker variability.
The speaker adaptation techniques can be grouped into three categories: feature-space, model-space, and feature augmentation based \cite{li2014overview,gales1998maximum,li2010comparison,saon2013speaker,swietojanski2016learning,wang2020speaker}. 
A common feature-space speaker adaptation technique is LIN \cite{seide2011feature,narayanan2014investigation,wang2019bridging}. It learns a speaker-specific linear transformation for the acoustic model input. 
LIN is simple and obtains high performance in various applications.
In particular, LIN reduces the word error rate substantially in robust ASR tasks, as shown in several works \cite{wang2018utterance,wang2018filter,wang2019enhanced}.

It is important to note that, unlike adaptation tasks, the LIN in this work is trained exclusively with the same training data as the original ST model, and no development or test data are exposed to the LIN. Each LIN is trained with its corresponding language-specific subset of training data. Experimental results show that LIN can be a simple and effective solution to the soft LID problem in language-agnostic many-to-one E2E ST. It can enhance the performance of the selected source language, without significantly compromising the quality of the other languages. Moreover, because LIN is initialized as an identity matrix, we can restore the original model by resetting the trained LIN layer to an identity matrix, guaranteeing that there is no degradation in the overall performance.

The remainder of this paper is organized as follows. We describe language-agnostic many-to-one E2E ST models and the proposed LIN method in Section \ref{sec:method}. Experimental setup is shown in Section \ref{sec:exp}. We analyze the evaluation results in Section \ref{sec:eval} and make a conclusion in Section \ref{sec:conc}.


\section{Method Description}
\label{sec:method}
\subsection{Language-Agnostic Many-to-One E2E ST}
\label{ssec:many_to_one}
E2E ST models can easily translate from multiple source languages to one target language using a single model. This is because that different source languages share the same audio representation space. To train such a model, we can simply mix the training data from all the source languages. In this way, we implicitly perform an LID detection task within the ST task. During inference, the model does not need any LID information for the source languages and can handle any input language with better code-switching support. E2E ST models are more user-friendly than conventional many-to-one ST systems, because they do not require the user to choose the source language from a list, which can be difficult when the speech contains code-switching. If the language selection from the user or the LID detection model is incorrect, the ST model might generate nonsensical output.
In addition to the above advantages, mixing different languages during training may also help each other and achieve higher overall performance.

Figure \ref{fig:comparison_lin} (a) shows a many-to-one ST model that can take three input languages and produce EN as the output language. This structure is used by both transducer-based models such as LAMASSU and AED-based models such as Whisper.




\subsection{Neural Transducer for Streaming Many-to-One E2E ST}
\label{ssec:transducer}
In this paper, we concentrate on many-to-one E2E ST models that employ neural transducers and refer to them as simplified versions of LAMASSU.These models are suitable for streaming scenarios and can also operate in offline mode with a large chunk size. We point out that our method is not restricted to neural transducer-based models, but can also be used for AED-based many-to-one E2E models.

A neural transducer consists of three components: an encoder, a prediction network, and a joint network. The encoder transforms audio features into hidden representations. The prediction network captures the dependency among output tokens. The joint network merges the outputs of the encoder and the prediction network. The key difference between neural transducer and AED models is that neural transducers operate at the frame level, while AED models operate at the token level.

Neural transducers have the advantage of performing streaming processing with per-frame granularity. We use the term Transformer-transducer (T-T) to refer to a neural transducer that employs a Transformer as its encoder. T-T has shown promising results in E2E ST \cite{xue2022large,wang2022lamassu,xue2023weakly}. To achieve streaming capability, we train the encoder with streaming masks and process the input in chunks \cite{xiechen2021tt}. The transducer loss accounts for all possible alignments of audio frames and output tokens, and thus automatically balances performance and latency for ST. Moreover, because the model can use the input features when computing the loss, it might achieve better convergence than AED models.


\subsection{Linear Input Network for Many-to-One ST}
\label{ssec:lin}



Language-agnostic many-to-one ST enhances user experiences significantly, but it has difficulty utilizing the LID information.
Most conventional methods for ST rely on source LID information and are hard-coded for specific languages. This breaks the language-agnostic property and can lead to nonsensical output when the source LID does not match the actual source language.

In this paper, we present LIN, a simple technique that is commonly applied in speaker adaptation, for the soft LID problem of language-agnostic many-to-one ST models. It can enhance the performance of the target language, while preserving the performance of the other languages. Specifically for the LAMASSU structure, we denote such method as LAMASSU-LIN.

We use a $D\times D$ LIN for a $D$-dimensional input audio features of a neural network. The LIN parameters are initialized as an identity matrix. During training, we fix all the other model parameters and only update the LIN weights using the training data of the specific source language. At test time, we pass all test utterances through the LIN. Since the LIN is tailored for the target language, the performance of that language may improve. On the other hand, the performances of the other input languages will be degraded. The degradation on the other languages is small because of the limited modeling power of LIN, which ensures that the model can still produce reasonable output when the LID specification is incorrect.
We note that applying LIN should generally improve the overall user experience, assuming that most of the traffic is from the specified language after specifying the input language. In addition, we initialize the LIN as an identity mapping. If LIN does not improve the model performance, we can switch back to identity mapping.

Note that LIN for soft LID and LIN for speaker adaptation are not the same. Speaker adaptation requires development or test data to adjust the model parameters. LIN for soft LID uses the training data to learn the model. This paper does not cover the use of development or test data for LIN-based soft LID adaptation, but such data could enhance the performance for the target language further.

Figure \ref{fig:comparison_lin} (b) illustrates the proposed soft LID module that uses LIN for many-to-one ST. In the figure, we train an LIN for language 2. The LIN handles the audio features of all languages during inference. Since the LIN is tailored for language 2, its output is expected to enhance the performance on language 2. However, the LIN might also impair the performance of the other languages, as indicated by the gray arrows.


\section{Experimental Setup}
\label{sec:exp}

\subsection{Dataset}
\label{ssec:data}
Our experiments involve 27 input languages: EN, Chinese (ZH), Portuguese (PT), Spanish (ES), Italian (IT), DE, French (FR), Japanese (JA), Russian (RU), Korean (KO), Polish (PL), Norwegian (NB), Hungarian (HU), Greek (EL), Czech (CS), Romanian (RO), Swedish (SV), Danish (DA), Finnish (FI), Dutch (NL), Slovenian (SL), Slovak (SK), Lithuanian (LT), Estonian (ET), Bulgarian (BG), Arabic (AR), and Hindi (HI). We train the 25-language baseline on the first 25 languages. Our model produces EN as the output.

We use 12 languages from the CoVoST2 \cite{wang2021covost} test set that are also supported by our models: DE, ES, ET, FR, IT, JA, NL, PT, RU, SL, SV, ZH. To measure the quality of our models, we use the bilingual evaluation understudy (BLEU) scores.

\subsection{Models}
\label{ssec:models}
This paper uses T-T models with a chunk size of 1 second for all experiments. We selected 1 second because it optimizes the trade-off between performance and latency. This is a simplified version of LAMASSU that only produces English output and has a shared encoder.

We train two LAMASSU-LIN models, one for JA input and the other for DE. They are denoted as LAMASSU-LIN-JA and LAMASSU-LIN-DE, respectively. For each of the two models, we append a 80$\times$80 trainable LIN to the 80-dimensional input feature. The LIN is defined as a nn.Linear layer in PyTorch with bias set to False. We then set the weight.data of the linear layer to a 80$\times$80 diagonal matrix for initialization. This ensures that the initial state of the LAMASSU-LIN model is identical to the original LAMASSU model. All the other parts of the model are initialized with the well-trained T-T model and their weights are fixed during training. We use the training data for the given input language to update the LIN layer. 

\subsection{Implementation Details}
\label{ssec:exp_details}
We train both LAMASSU-LIN-JA and LAMASSU-LIN-DE for 6400000 steps using only the respective language and a custom Noam decay optimizer with a peak learning rate of 5e-5. The warmup step is 800000 and the base value is 512. We use PyTorch DDP with DeepSpeed on 32 32G GPUs. The fp16 argument is True with loss scale 0 and minimum loss scale 1. The batch size is 400000 for the product of frame length and label length. We exclude utterances exceeding 30 seconds in audio length and those with a token count under 3 or over 230. The number of buffered utterances is set to 10000 to speed up data I/O.

\section{Evaluation Results}
\label{sec:eval}

\subsection{The Impact of More Languages}
\label{ssec:eval_more}

\begin{table}[htb]
  \caption{Comparisons of BLEU scores between the 25-language baseline and the 27-language baseline. We do not include the AR result in the table since it is not one of the languages in the 25-language baseline, but the BLEU score on the AR test set of CoVoST2 can reach 45.3 using LAMASSU-LIN-DE.} 
  \label{tab:results_25_27}
  \centering
  \begin{tabular}{l c c c c}
    \toprule
    languages & 25-language \cite{xue2023weakly} & 27-language \\
    \midrule
    DE & 34.0 & 34.8 \\
    ES & 34.7 & 35.4 \\
    ET & 17.9 & 19.4 \\
    FR & 33.0 & 34.0 \\
    IT & 33.4 & 34.2 \\
    JA & 21.6 & 23.2 \\
    NL & 38.5 & 40.2 \\
    PT & 44.7 & 46.0 \\
    RU & 39.8 & 40.9 \\
    SL & 22.4 & 23.5 \\
    SV & 37.1 & 38.5 \\
    ZH & 18.0 & 18.2 \\
    \midrule
    Avg. & 31.3 & 32.4 \\
    \bottomrule
  \end{tabular}
\end{table}

\begin{table*}[htb]
  \caption{Comparisons of BLEU scores among the 27-language baseline, the LAMASSU-LIN model for JA (LAMASSU-LIN-JA) and that for DE (LAMASSU-LIN-DE). The 99\% traffic on DE result of the 27-language baseline is 34.77, and that of LAMASSU-LIN-DE is 34.83. Therefore, we only highlight the LAMASSU-LIN-DE result.} 
  \label{tab:results_ja_de}
  \centering
  \begin{tabular}{l c c c}
    \toprule
    languages & 27-language & LAMASSU-LIN-JA & LAMASSU-LIN-DE \\
    \midrule
    DE & 34.8 & 32.1 & \textbf{34.9} \\
    ES & 35.4 & 33.7 & 34.3 \\
    ET & 19.4 & 18.1 & 19.0 \\
    FR & 34.0 & 32.1 & 32.3 \\
    IT & 34.2 & 32.5 & 32.8 \\
    JA & 23.2 & \textbf{24.0} & 22.7 \\
    NL & 40.2 & 38.3 & 38.9 \\
    PT & 46.0 & 43.7 & 43.9 \\
    RU & 40.9 & 38.9 & 39.6 \\
    SL & 23.5 & 19.3 & 23.2 \\
    SV & 38.5 & 36.1 & 37.4 \\
    ZH & 18.2 & 16.7 & 17.0 \\
    \midrule
    Avg. & 32.4 & 30.4 & 31.3 \\
    Weighted Avg. w/ 99\% traffic from JA & 23.3 & \textbf{24.1} & 22.8 \\
    Weighted Avg. w/ 99\% traffic from DE & {34.8} & 32.1 & \textbf{34.8} \\
    \bottomrule
  \end{tabular}
\end{table*}

Table \ref{tab:results_25_27} shows the comparison between the 25-language baseline model and the 27-language baseline model. By adding two more languages, AR and HI, the average BLEU score on the 12 languages increases by 1.1. The BLEU score for each of the 12 languages also improves. This suggests that transducer-based many-to-one E2E ST models can handle more languages effectively. It indicates that the advantage of having more language directions is greater than the interference of different languages in many-to-one E2E ST. A possible explanation is that the model learns more acoustic variations with more training data.
Indeed, we can scale up to more than 70 languages with competitive results.


\subsection{LAMASSU-LIN}
\label{ssec:eval_lid}
Table \ref{tab:results_ja_de} compares the 27-language baseline and the proposed soft LID method for JA and DE. The BLEU score of JA increases by 0.8 with LAMASSU-LIN-JA, while the other languages decrease. However, the average BLEU score of LAMASSU-LIN-JA remains above 30. The BLEU score of DE rises slightly from 34.8 to 34.9 with LAMASSU-LIN-DE. The other languages also drop, but the average score of LAMASSU-LIN-DE is higher than LAMASSU-LIN-JA. This demonstrates the soft LID capability, which balances the improvement of the target language and the deterioration of the other languages. For DE, the smaller improvement leads to smaller degradation as well. This soft LID capability distinguishes our method from most existing methods, which can produce completely irrelevant outputs if the LID is incorrect. We believe this feature stems from the limited modeling power of the trainable part, which prevents it from excessively skewing the model.

We simulate real-world scenarios where 99\% of the traffic is from the specified languages and compare the weighted average BLEU scores. These comparisons assume that the source LID information, either given by the user or estimated from an LID detection model, is mostly correct. Note that for hard-coded LID methods, even though the chance of using a wrong LID is also low under this assumption, the outputs when the LID is wrong are completely irrelevant and confusing to users, resulting in a huge impact on user experiences.
In a scenario where 99\% of the traffic is from JA and the remaining 1\% is equally distributed among the other 11 languages, we compute the weighted average BLEU scores. Compared with the 27-language baseline result of 23.3, LAMASSU-LIN-JA achieved 24.1 BLEU score, surpassing the baseline by 0.8. We also present the results when 99\% of the traffic is from DE and the remaining 1\% is equally distributed among the other 11 languages in the last row. Both the 27-language baseline and LAMASSU-LIN-DE obtained 34.8 BLEU score. Note that the same BLEU score values are due to rounding. The more exact values are 34.77 for the 27-language baseline and 34.83 for LAMASSU-LIN-DE. Therefore, LAMASSU-LIN-DE performs the best among the three models in this scenario.



\section{Concluding Remarks}
\label{sec:conc}
We introduce the soft LID problem for language-agnostic many-to-one E2E ST models and propose a simple method denoted as LAMASSU-LIN. LAMASSU-LIN applies LIN, a common method in speaker adaptation, to the soft LID problem of a transducer-based E2E ST model. Note that different from adaptation tasks, we train the LIN only using the language-specific subset inside the training set, without involving the development or test sets.
Our approach enhances the ST performance for the specified language. At the same time, despite minor decreases in BLEU scores, the proposed approach preserves most of the ST quality for the other languages. When 99\% of the traffic is from the specified language, our methods surpass the baselines. 

One known limitation of this work is that for some languages, LIN may yield worse results for the specified language. This may be because the specified language benefits from multilingual training using other languages. This is also the reason why the capability of being reset to identity mapping is an important feature for the soft LID problem. In addition, the influence between different languages in a multilingual ST model is still an open research area that may need to be studied.

In this paper, we only explore LIN, but we believe there are many other methods that can achieve similar goals and possibly better outcomes for the soft LID problem. Moreover, these methods, including LIN, may be applied to not only transducer-based streaming many-to-one E2E ST models, but also AED models such as Whisper and other multimodal generative models such as GPT-4o, which we may investigate in the future.

\bibliographystyle{IEEEtran}
\bibliography{mybib}

\begin{thebibliography}{10}
\providecommand{\url}[1]{#1}
\csname url@samestyle\endcsname
\providecommand{\newblock}{\relax}
\providecommand{\bibinfo}[2]{#2}
\providecommand{\BIBentrySTDinterwordspacing}{\spaceskip=0pt\relax}
\providecommand{\BIBentryALTinterwordstretchfactor}{4}
\providecommand{\BIBentryALTinterwordspacing}{\spaceskip=\fontdimen2\font plus
\BIBentryALTinterwordstretchfactor\fontdimen3\font minus \fontdimen4\font\relax}
\providecommand{\BIBforeignlanguage}[2]{{%
\expandafter\ifx\csname l@#1\endcsname\relax
\typeout{** WARNING: IEEEtran.bst: No hyphenation pattern has been}%
\typeout{** loaded for the language `#1'. Using the pattern for}%
\typeout{** the default language instead.}%
\else
\language=\csname l@#1\endcsname
\fi
#2}}
\providecommand{\BIBdecl}{\relax}
\BIBdecl

\bibitem{li2022recent}
J.~Li, ``Recent advances in end-to-end automatic speech recognition,'' \emph{APSIPA Transactions on Signal and Information Processing}, vol.~11, no.~1, 2022.

\bibitem{vila2018end}
L.~C. Vila, C.~Escolano, J.~A. Fonollosa, and M.~R. Costa-Jussa, ``End-to-end speech translation with the transformer.'' in \emph{Proc. of INTERSPEECH}, 2018, pp. 60--63.

\bibitem{sperber2020speech}
M.~Sperber and M.~Paulik, ``Speech translation and the end-to-end promise: Taking stock of where we are,'' in \emph{Proc. of ACL}, 2020, pp. 7409--7421.

\bibitem{wang2019token}
P.~Wang, J.~Cui, C.~Weng, and D.~Yu, ``Token-wise training for attention based end-to-end speech recognition,'' in \emph{Proc. of ICASSP}, 2019, pp. 6276--6280.

\bibitem{wang2019large}
------, ``Large margin training for attention based end-to-end speech recognition,'' in \emph{Proc. of INTERSPEECH}, 2019, pp. 246--250.

\bibitem{wang21t_interspeech}
P.~Wang, T.~N. Sainath, and R.~J. Weiss, ``Multitask training with text data for end-to-end speech recognition,'' in \emph{Proc. of Interspeech}, 2021, pp. 2566--2570.

\bibitem{Graves-RNNSeqTransduction}
A.~Graves, ``Sequence transduction with recurrent neural networks,'' \emph{arXiv preprint arXiv:1211.3711}, 2012.

\bibitem{prabhavalkar2017asr}
R.~Prabhavalkar, K.~Rao, T.~N. Sainath, B.~Li, L.~Johnson, and N.~Jaitly, ``A comparison of sequence-to-sequence models for speech recognition,'' in \emph{Proc. of INTERSPEECH}, 2017, pp. 939--943.

\bibitem{sainath2020asr}
T.~N. Sainath, Y.~He, B.~Li, A.~Narayanan, R.~Pang, A.~Bruguier, S.-Y. Chang, W.~Li, R.~Alvarez, Z.~Chen, and et~al, ``A streaming on-device end-to-end model surpassing server-side conventional model quality and latency,'' in \emph{Proc. of ICASSP}, 2020, pp. 6059--6003.

\bibitem{li2020asr}
J.~Li, R.~Zhao, Z.~Meng, Y.~Liu, W.~Wei, S.~Parthasarathy, V.~Mazalov, Z.~Wang, L.~He, S.~Zhao, and et~al, ``Developing rnnt models surpassing high-performance hybrid models with customization capability,'' in \emph{Proc. of INTERSPEECH}, 2020, pp. 3590--3594.

\bibitem{saon2021asr}
G.~Saon, Z.~Tüske, D.~Bolanos, and B.~Kingsbury, ``Advancing rnn transducer technology for speech recognition,'' in \emph{Proc. of ICASSP}, 2021, pp. 5654--5658.

\bibitem{Berard2016ST}
A.~Berard, O.~Pietquin, C.~Servan, and L.~Besacier, ``Listen and translate: A proof of concept for end-to-end speech-to-text translation,'' in \emph{NIPS Workshop on End-to-end Learning for Speech and Audio Processing}, 2016.

\bibitem{weiss2017sequence}
R.~J. Weiss, J.~Chorowski, N.~Jaitly, Y.~Wu, and Z.~Chen, ``Sequence-to-sequence models can directly translate foreign speech,'' in \emph{Proc. of INTERSPEECH}, 2017, pp. 2625--2629.

\bibitem{Berard2018ST}
A.~B{\'e}rard, L.~Besacier, A.~C. Kocabiyikoglu, and O.~Pietquin, ``End-to-end automatic speech translation of audiobooks,'' in \emph{Proc. of ICASSP}.\hskip 1em plus 0.5em minus 0.4em\relax IEEE, 2018, pp. 6224--6228.

\bibitem{radford2023robust}
A.~Radford, J.~W. Kim, T.~Xu, G.~Brockman, C.~McLeavey, and I.~Sutskever, ``Robust speech recognition via large-scale weak supervision,'' in \emph{International Conference on Machine Learning}.\hskip 1em plus 0.5em minus 0.4em\relax PMLR, 2023, pp. 28\,492--28\,518.

\bibitem{papi2023token}
S.~Papi, P.~Wang, J.~Chen, J.~Xue, J.~Li, and Y.~Gaur, ``Token-level serialized output training for joint streaming asr and st leveraging textual alignments,'' in \emph{2023 IEEE Automatic Speech Recognition and Understanding Workshop (ASRU)}.\hskip 1em plus 0.5em minus 0.4em\relax IEEE, 2023, pp. 1--8.

\bibitem{yang2023diarist}
M.~Yang, N.~Kanda, X.~Wang, J.~Chen, P.~Wang, J.~Xue, J.~Li, and T.~Yoshioka, ``Diarist: Streaming speech translation with speaker diarization,'' \emph{arXiv preprint arXiv:2309.08007}, 2023.

\bibitem{papi2023leveraging}
S.~Papi, P.~Wang, J.~Chen, J.~Xue, N.~Kanda, J.~Li, and Y.~Gaur, ``Leveraging timestamp information for serialized joint streaming recognition and translation,'' \emph{arXiv preprint arXiv:2310.14806}, 2023.

\bibitem{xue2022large}
J.~Xue, P.~Wang, J.~Li, M.~Post, and Y.~Gaur, ``Large-scale streaming end-to-end speech translation with neural transducers,'' in \emph{Proc. of INTERSPEECH}, 2022, pp. 3263--3267.

\bibitem{wang2022lamassu}
P.~Wang, E.~Sun, J.~Xue, Y.~Wu, L.~Zhou, Y.~Gaur, S.~Liu, and J.~Li, ``{LAMASSU}: Streaming language-agnostic multilingual speech recognition and translation using neural transducers,'' in \emph{Proc. of INTERSPEECH}, 2023, pp. 57--61.

\bibitem{liu2021caat}
D.~Liu, M.~Du, X.~Li, Y.~Li, and E.~Chen, ``Cross attention augmented transducer networks for simultaneous translation,'' in \emph{Proceedings of EMNLP}, 2021, pp. 39--55.

\bibitem{tang2023hybrid}
Y.~Tang, A.~Y. Sun, H.~Inaguma, X.~Chen, N.~Dong, X.~Ma, P.~D. Tomasello, and J.~Pino, ``Hybrid transducer and attention based encoder-decoder modeling for speech-to-text tasks,'' \emph{arXiv preprint arXiv:2305.03101}, 2023.

\bibitem{xue2023weakly}
J.~Xue, P.~Wang, J.~Li, and E.~Sun, ``A weakly-supervised streaming multilingual speech model with truly zero-shot capability,'' in \emph{2023 IEEE Automatic Speech Recognition and Understanding Workshop (ASRU)}.\hskip 1em plus 0.5em minus 0.4em\relax IEEE, 2023, pp. 1--7.

\bibitem{li2014overview}
J.~Li, L.~Deng, Y.~Gong, and R.~Haeb-Umbach, ``An overview of noise-robust automatic speech recognition,'' \emph{IEEE/ACM Transactions on Audio, Speech, and Language Processing}, vol.~22, no.~4, pp. 745--777, 2014.

\bibitem{gales1998maximum}
M.~Gales, ``Maximum likelihood linear transformations for hmm-based speech recognition,'' \emph{Computer speech \& language}, vol.~12, no.~2, pp. 75--98, 1998.

\bibitem{li2010comparison}
B.~Li and K.~Sim, ``Comparison of discriminative input and output transformations for speaker adaptation in the hybrid dnn/hmm systems,'' in \emph{Eleventh Annual Conference of the International Speech Communication Association}, 2010.

\bibitem{saon2013speaker}
G.~Saon, H.~Soltau, D.~Nahamoo, and M.~Picheny, ``Speaker adaptation of neural network acoustic models using i-vectors.'' in \emph{ASRU}, 2013, pp. 55--59.

\bibitem{swietojanski2016learning}
P.~Swietojanski, J.~Li, and S.~Renals, ``Learning hidden unit contributions for unsupervised acoustic model adaptation,'' \emph{IEEE/ACM Transactions on Audio, Speech, and Language Processing}, vol.~24, no.~8, pp. 1450--1463, 2016.

\bibitem{wang2020speaker}
P.~Wang, Z.~Chen, D.~L. Wang, J.~Li, and Y.~Gong, ``Speaker separation using speaker inventories and estimated speech,'' \emph{IEEE/ACM TASLP}, vol.~29, pp. 537--546, 2020.

\bibitem{seide2011feature}
F.~Seide, G.~Li, X.~Chen, and D.~Yu, ``Feature engineering in context-dependent deep neural networks for conversational speech transcription,'' in \emph{Automatic Speech Recognition and Understanding (ASRU), 2011 IEEE Workshop on}.\hskip 1em plus 0.5em minus 0.4em\relax IEEE, 2011, pp. 24--29.

\bibitem{narayanan2014investigation}
A.~Narayanan and D.~L. Wang, ``Investigation of speech separation as a front-end for noise robust speech recognition,'' \emph{IEEE/ACM Transactions on Audio, Speech and Language Processing (TASLP)}, vol.~22, no.~4, pp. 826--835, 2014.

\bibitem{wang2019bridging}
P.~Wang, K.~Tan \emph{et~al.}, ``Bridging the gap between monaural speech enhancement and recognition with distortion-independent acoustic modeling,'' \emph{IEEE/ACM TASLP}, vol.~28, pp. 39--48, 2019.

\bibitem{wang2018utterance}
P.~Wang and D.~L. Wang, ``Utterance-wise recurrent dropout and iterative speaker adaptation for robust monaural speech recognition,'' in \emph{Proc. of ICASSP}.\hskip 1em plus 0.5em minus 0.4em\relax IEEE, 2018, pp. 4814--4818.

\bibitem{wang2018filter}
P.~Wang and D.-L. Wang, ``Filter-and-convolve: A cnn based multichannel complex concatenation acoustic model,'' in \emph{2018 IEEE International Conference on Acoustics, Speech and Signal Processing (ICASSP)}.\hskip 1em plus 0.5em minus 0.4em\relax IEEE, 2018, pp. 5564--5568.

\bibitem{wang2019enhanced}
P.~Wang and D.~L. Wang, ``Enhanced spectral features for distortion-independent acoustic modeling,'' in \emph{Proc. of INTERSPEECH}, 2019, pp. 476--480.

\bibitem{xiechen2021tt}
X.~Chen, Y.~Wu, Z.~Wang, S.~Liu, and J.~Li, ``Developing real-time streaming transformer transducer for speech recognition on large-scale dataset,'' in \emph{Proc. of ICASSP}.\hskip 1em plus 0.5em minus 0.4em\relax IEEE, 2021, pp. 5904--5908.

\bibitem{wang2021covost}
C.~Wang, A.~Wu, J.~Gu, and J.~Pino, ``Covost 2 and massively multilingual speech translation.'' in \emph{Proc. of INTERSPEECH}, 2021, pp. 2247--2251.

\end{thebibliography}

\end{document}